\documentclass[letterpaper, 10 pt, conference]{ieeeconf}  

\IEEEoverridecommandlockouts                              
\usepackage{multirow}
\usepackage{booktabs}
\usepackage{amsmath}
\usepackage{amssymb}
\usepackage{float}
\usepackage[linesnumbered,ruled]{algorithm2e}
\usepackage{cite}
\usepackage{graphicx}
\usepackage{subcaption}
\usepackage{xcolor}
\usepackage{makecell}
\usepackage{bm}
\usepackage{tabularx}
\usepackage{threeparttable}
\definecolor{darkyellow}{RGB}{225, 153, 0}
\definecolor{darkgreen}{RGB}{0, 100, 0}
\usepackage[colorlinks,linkcolor=red,anchorcolor=blue,citecolor=green]{hyperref}
\title{\LARGE \bf
Semantic Region Aware Autonomous Exploration for Multi-Type Map Construction in Unknown Indoor Environments
}

\author{Jianfang Mao$^\dagger$
\thanks{Jianfang Mao is with the Chongqing University, Chongqing 400044, China.}
\thanks{$\dagger$ Jianfang Mao is the first author.}%
}

\begin{document}

\maketitle

\begin{abstract}

Mainstream autonomous exploration methods usually perform excessively-repeated explorations for the same region, leading to long exploration time and exploration trajectory in complex scenes. 
To handle this issue, we propose a novel semantic region aware autonomous exploration method, the core idea of which is considering the information of semantic regions to optimize the autonomous navigation strategy.    
Our method enables the mobile robot to fully explore the current semantic region before moving to the next region, contributing to avoid excessively-repeated explorations and accelerate the exploration speed. 
In addition, compared with existing autonomous exploration methods that usually construct the single-type map, our method allows to construct four types of maps including point cloud map, occupancy grid map, topological map, and semantic map. 
The experiment results demonstrate that our method achieves the highest \textbf{50.7\%} exploration time reduction and \textbf{48.1\%} exploration trajectory length reduction while maintaining \textbf{\(>\)98\%} exploration rate when comparing with the classical RRT (Rapid-exploration Random Tree) based autonomous exploration method.

\end{abstract}

\section{Introduction}

Map construction via autonomous exploration is a task that a robot moves in an unknown environment and synchronously construct the map of the environment, which is significant for robotic systems \cite{r50,r51,r52,r53,r54}. The frontier-based autonomous exploration mechanism is widely used in existing methods \cite{r16,r55,r41,r56,r42}. 
The seminal work \cite{r16}
firstly detects the frontier between the unknown region (i.e., the region that has not been explored) and the known region (i.e., the region that has been explored) using the laser scanner. Then, some candidate frontier points are generated based on the frontier. Subsequently, the nearest frontier point among the candidate frontier points is selected robot's moving goal. The above steps are repeated to finally realize the exploration of the whole environment.Based on the frontier-based mechanism, the NBV (Next-Best-View) based exploration mechanism optimizes the candidate frontier point evaluation function
by considering the information gain \cite{r17}, path cost \cite{r32}, and other factors \cite{r18} of frontier points. Apart from the above exploration mechanics, some methods \cite{r35,r5,r4,r6,r40,r57,r58} propose the sample-based mechanism to perform the exploration.  

\begin{figure}[!ht]
  \vspace{6pt} 
   \centering
  \includegraphics[width=0.48\textwidth]{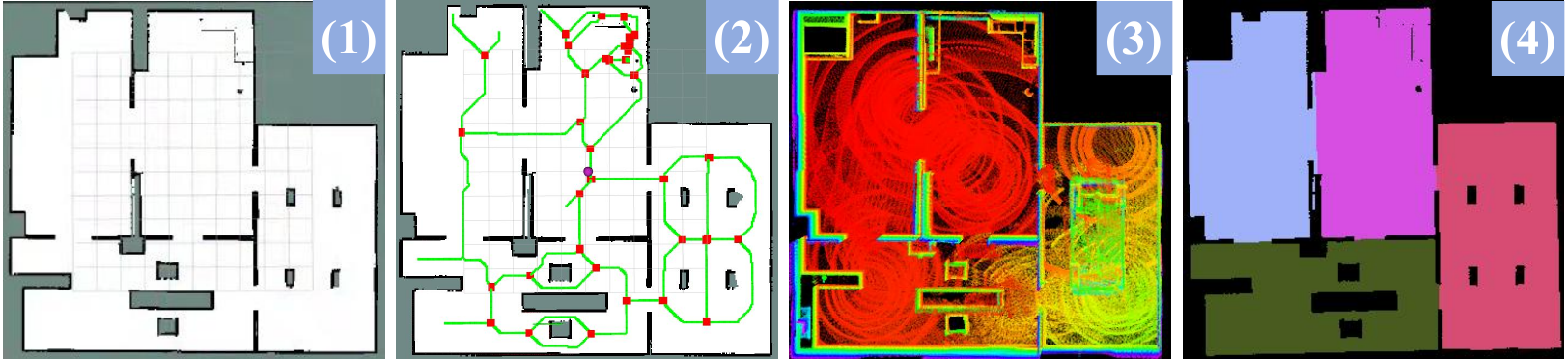}
  \caption{Four types of maps constructed by our semantic region aware autonomous exploration method. (1) 2D occupancy grid map, (2) topological map, (3) 3D point cloud map, and (4) semantic map.}
  \label{fig:out put maps}
\end{figure} 

These methods have significantly pushed forward the research of autonomous exploration, but present two insufficiencies. \textit{1) Existing autonomous exploration methods do not simultaneously generate the rich types of maps}. Richer types of maps could support a wider range of downstream tasks and applications. For example, the occupancy grid and topological maps aid in path planning \cite{r44,r61,r62}, the point cloud map is key for localization and 3D detection \cite{r43,r63,r65}, and the semantic map can enhance human-robot interaction by providing the scene-level understanding. Some works only generate the single-type map \cite{r16,r40,r34}. Although some studies attempt to generate multi-type maps, 
such as the occupancy grid map and the topological map in \cite{r10,r8,r28}, and the point cloud map and the topological map in \cite{r14}, the map types are still not rich. If the robot cannot generate rich map types
during the autonomous exploration, it often requires re-exploration or manual intervention when these maps are needed for subsequent tasks.

\textit{2) Existing autonomous exploration methods
usually execute excessively-repeated explorations for the same region.} When reproducting existing methods, we find it is a common case that a robot moves to the next region when the current region has not been fully explored, which easily generates repeated exploration trajectories and significantly affects the exploration efficiency. We analyze the reasons are two-fold.
First, due to the randomness of candidate frontier points, it is difficult to stably guarantee that the next best viewpoint goal is always inside the current region before it is fully explored. Second, the frontier point evaluation function does not consider the semantic region information of environment when determining the next best viewpoint goal, so it is easy to select the frontier point (closer to other bigger unknown space) as the best viewpoint. 

To handle the above two insufficiencies, this paper proposes a semantic region aware autonomous exploration method, which encourages a mobile robot to fully explore the current region before moving to the next region. The proposed method achieves faster speed by avoiding a robot to come back again to explore the previously-unexplored space in the current region, which is implemented by proposing a new frontier point generation mechanism and a new frontier point evaluation function that take the semantic region information of environments into the consideration. In addition, the proposed autonomous exploration method could generate four types of maps (i.e., the 2D occupancy grid map, the 3D point cloud map, the topological map, and the semantic map shown in Fig.~\ref{fig:out put maps})~while maintaining the real-time exploration at the same time.

In the experiments, our method is compared with original RRT \cite{r6}, TOPO \cite{r28}, Improved RRT and MMPF (proposed in \cite{r22}) in three simulated environments. Compared to classical RRT, our method achieves \textbf{50.7\%} exploration time reduction and \textbf{48.1\%} exploration trajectory length reduction when maintaining \textbf{\(>\)98\%} exploration rate. 
We also compare the map types of our method with that of existing methods.
In addition, the storage size and update time of different types of maps are analyzed. 

The contributions of this paper are as follows:
\begin{itemize}
\item This paper proposes a semantic region aware autonomous exploration method, which is able to significantly improve the exploration efficiency. 


\item The proposed autonomous exploration method allows to simultaneously construct four types of maps in unknown indoor environments.
\end{itemize}

\section{Related Work}

\subsection{ Autonomous Exploration Strategy }


Widely used robotic exploration strategies include the frontier-based mechanism \cite{r16,r55,r41,r56,r42} and the NBV-based mechanism \cite{r17,r32,r18}. However, both strategies face challenges with high computational costs in complex, large environments. To address this issue, subsequent researchers proposed exploration methods based on the sample-based mechanisms to reduce computational costs.

The sample-based mechanism exploration strategies mainly include the RRT family \cite{r35,r5,r4,r6,r40} and the Random Roadmap family \cite{r46,r47,r66}. RRT \cite{r6} achieves tree growth through continuous random sampling in the map. When the tree reaches the frontiers, it generates a goal. In the work \cite{r35}, to improve the sampling efficiency of the original RRT, the idea of a disjointed tree was proposed. Random roadmaps can also be used for path planning. Wang et al. \cite{r47} narrowed the search area from the entire map to the free space and extended the graph-based roadmap, thus supporting multiple queries and facilitating path planning.

Although sample-based mechanism exploration strategies have made significant improvements in computational efficiency, they can lead to the problem of excessive repetition in exploration due to the randomness of sample point selection. By incorporating semantic maps, we introduced semantic region aware points into the random sampling process. Additionally, in the frontier evaluation function, we integrated semantic region information alongside information gain and path cost, effectively addressing this issue.

\subsection{ Hybrid Mapping System}

Hybrid mapping systems \cite{r64} are widely studied to provide comprehensive information for robot tasks. In this section, the related work is classified according to the structures created for hybrid mapping. The first group pertains to works that generated 2D occupancy grid and topological map \cite{r10}, \cite{r8}. In Zhang's work \cite{r8}, the idea of Voronoi diagrams was utilized to construct a topological map after building a occupancy grid map. In \cite{r10}, occupancy grid and topological maps were constructed in real time, and priority values were assigned to topological nodes according to their environment regions, so as to realize the graph exploration algorithm based on the priority of topological nodes.

The second group consists of a hybrid map composed of 3D point clouds and topological maps \cite{r12}, \cite{r11}. In the work \cite{r12}, a topological representation of free space maps to navigation graphs and convex voxel clusters was proposed. To improve the efficiency of global path planning, Xue etc. \cite{r11} built the topological map using both map points and trajectories of visual SLAM. The first two groups share a common problem: these hybrid maps cannot help the robot understand the environment like humans. If the target is obstructed by obstacles, the robot may not realize it even when it is close to the target. Adding semantic information to the region can avoid this problem. Following this, the third group aims to add semantic information to the hybrid map. The third group added semantic information to the hybrid metric and topological maps \cite{r14,r27,r15,r9,r13}. In \cite{r13}, each node of the topological map contains a set of images from the region as semantic information, along with added metric information. In \cite{r14}, the topological global representation and 3D dense submaps were maintained as a hybrid global map, which could be built by using a standard CPU, reducing the computational resources required. In \cite{r15}, both unoccupied and occupied areas were characterized by voronoi diagrams, with recognized and classified objects from camera views placed in the topological nodes.

Although hybrid mapping systems have been studied extensively before, the types of maps that can be constructed simultaneously by a mobile robot have not been comprehensive enough. Thus, this work aims to fill this gap.

\begin{figure*}[ht]
  \vspace{2pt} 
  \centering
  \includegraphics[width=1\textwidth]{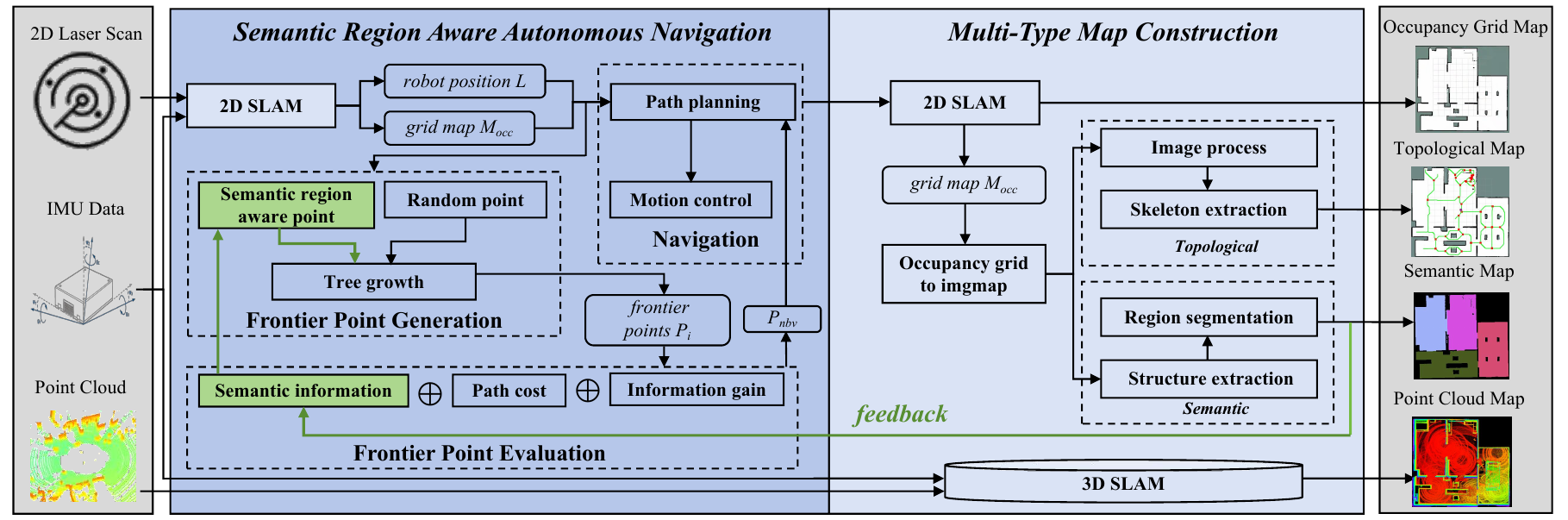}
 \caption{The overview of our method. The semantic region aware parts are shaded in green. }
  \vspace{-18pt}
  \label{fig:system_overview}
\end{figure*}

\section{Method}
\subsection{Preliminaries: RRT-Based Exploration }

RRT-based exploration \cite{r6} is a classical autonomous exploration method. 
An environment is composed of known space ($\bm{S}_{\textit{know}}$) which has been explored and unknown space ($\bm{S}_{\textit{unkn}}$) which has not been explored. 
The goal of autonomous exploration is to explore $\bm{S}_{\textit{unkn}}$. To this end, the method firstly generates random points $\bm{P}_\textit{rad}$. Then, based on $\bm{P}_\textit{rad}$ and the robot's initial location $\bm{L}_{\textit{init}}$, a tree branch originating from $\bm{L}_{\textit{init}}$ is growing to cover $\bm{P}_\textit{rad}$. Frontier points $\bm{P}_\textit{i}$ are computed based on the tree branches. Coarsely, $\bm{P}_\textit{i}$ are points on the tree branches, and $\bm{P}_\textit{i}$ also locate on the frontier (i.e., the boundary between $\bm{S}_{\textit{unkn}}$ and $\bm{S}_{\textit{know}}$), thus $\bm{P}_\textit{i}$ are the guidance of the moving direction towards $\bm{S}_{\textit{unkn}}$. Finally, the next best viewpoint goal $\bm{P}_{\textit{nbv}}$ is continuously selected from $\bm{P}_\textit{i}$ according to the frontier point evaluation function, and $\bm{P}_{\textit{nbv}}$ guides the robot to navigate to unknown space to realize autonomous exploration.

\subsection{Overview}

As shown in Fig.~\ref{fig:system_overview}, taking 2D laser scan, IMU data, and point cloud data as input, our method outputs four types of maps through two key modules, namely the semantic region aware autonomous navigation module and multi-type map construction module.
In the semantic region aware autonomous navigation module, 
frontier point generation mechanism outputs frontier points $\bm{P}_\textit{i}$ based on the 2D occupancy grid map $\bm{M}_{\textit{occ}}$ and the robot position $\bm{L}$. 
$\bm{P}_\textit{i}$ are then provided to the frontier point evaluation function, which outputs the robot's next best viewpoint goal $\bm{P}_{\textit{nbv}}$. Path planning is conducted in $\bm{M}_{\textit{occ}}$ based on $\bm{P}_{\textit{nbv}}$ and $\bm{L}$, guiding the robot to move to $\bm{P}_{\textit{nbv}}$. In the multi-type map construction module, four types of maps are constructed and updated.   

\subsection{Semantic Region Aware Autonomous Navigation}

Frontier points $\bm{P}_\textit{i}$ locate on the frontier (i.e., the boundary between $\bm{S}_{\textit{unkn}}$ and $\bm{S}_{\textit{know}}$),
thus they are important signals to guide the robot to explore $\bm{S}_{\textit{unkn}}$.
Since frontier point generation in RRT-based exploration relies on random sampling, the robot's exploration behavior easily result in excessively-repeated explorations for the same region. To address this issue, we propose a semantic region aware frontier point generation mechanism and semantic region aware frontier point evaluation function. 

\subsubsection{Semantic region aware frontier point generation mechanism}
\begin{figure}[ht]
    \vspace{-5pt}
    \centering
    \includegraphics[width=0.47\textwidth]{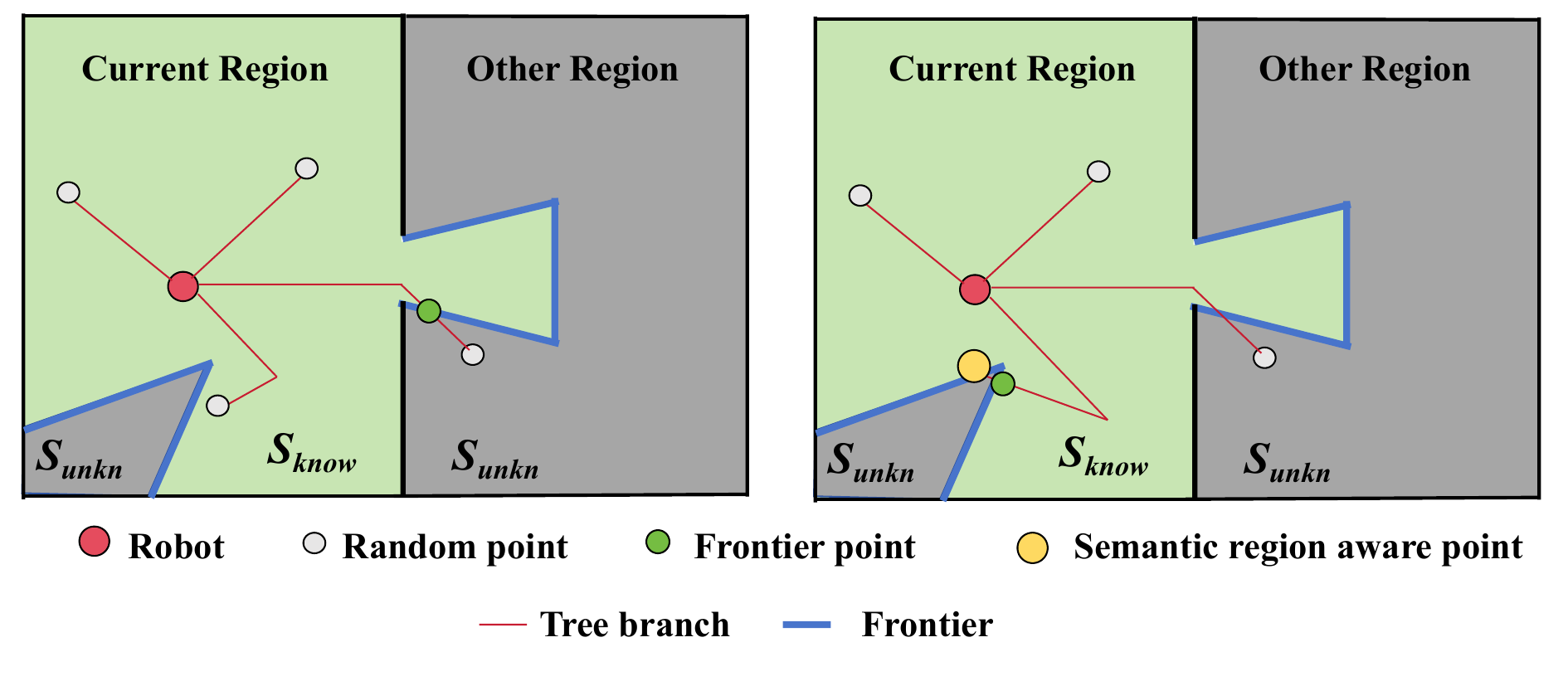}
    \label{fig:side:a1}
    \vspace{-10pt}
  \caption{Comparison of original frontier point generation (left) and semantic region aware frontier point generation (right).} 
  \label{fig:Comparison of frontier point}
 \vspace{-13pt}
\end{figure}

As shown in Fig.~\ref{fig:Comparison of frontier point}, since the original frontier point generation mechanism does not consider regional semantics, the frontier point near to the bigger unknown space is easily selected as the next-to-move point, leading to that the robot needs to come back again to explore the smaller unknown space in the current region. In big and complex environments, excessively-repeated explorations occur frequently.

To alleviate the excessively-repeated explorations, we firstly introduce the semantic region aware point $\bm{P}_\textit{sem}$ ( \textcolor{darkyellow}{yellow point} in Fig.~\ref{fig:Comparison of frontier point}), which meets the conditions that $\bm{P}_\textit{sem}$ is within the current semantic region and locates on the frontier. 
In conventional methods, the tree branch is growing based on random points $\bm{P}_\textit{rad}$ ( \textcolor{gray}{grey points} in Fig.~\ref{fig:Comparison of frontier point}). 
In our method, the tree branch is growing based on both $\bm{P}_\textit{rad}$ and $\bm{P}_\textit{sem}$. We propose a dynamic probability mechanism to select the sampling point $\bm{P}_{\textit{sap}}$, the function of $\bm{P}_{\textit{sap}}$ is to control the growing trend of tree branch. $\bm{P}_{\textit{sap}}$ is formulated as follow.
\begin{equation}
\label{eq:1}
\left\{
\begin{aligned}
    &  p(\bm{P_{\textit{sap}}} = \bm{P}_\textit{rad}) = \frac{1}{1 + k \cdot t}\\
    &  p(\bm{P_{\textit{sap}}} = \bm{P}_\textit{sem}) = \frac{k \cdot t}{1 + k \cdot t}
\end{aligned}
\right.
\end{equation}
where $p(\bm{P_{\textit{sap}}}=\bm{P}_\textit{rad})$ denotes the probability that $\bm{P}_\textit{rad}$ is selected as $\bm{P}_{\textit{sap}}$, $p(\bm{P_{\textit{sap}}}=\bm{P}_\textit{sem})$ denotes the probability that $\bm{P}_\textit{sem}$ is selected as $\bm{P}_{\textit{sap}}$, $t$ is a dynamic value signalling the exploration time in the current semantic region, and k controls the extent to which $t$ affects the probability. We note that the selection of $\bm{P}_\textit{sem}$ is based on semantic map, as the green feedback line in Fig.~\ref{fig:system_overview}. The process of generating the semantic map will be detailed in the multi-type map construction section.  

The small $t$ demonstrates the robot just begins to explore the current region, thus the tree branch grows based on $\bm{P}_\textit{rad}$.
With the increasing of $t$, $p(\bm{P_{\textit{sap}}} = \bm{P}_\textit{sem})$ becomes larger, which encourages the tree branch to grow to the unknown space in the current semantic region, prioritizing the robot to explore the current region rather than other regions.



After obtaining $\bm{P_{\textit{sap}}}$ by Eq.~\ref{eq:1}, the tree branch is determined. With the tree branch, a set of frontier points $\bm{F}$ are generated by judging whether the tree branch crosses with $\bm{S}_{\textit{unkn}}$. $\bm{F}$ is denoted as follow.
\begin{equation}
\begin{aligned}
    \bm{F} = \left\{ \bm{P}_\textit{i} ~ \lvert~ \textit{i} = 1,2,...,n \right\}
\end{aligned}
\label{eq:front_points}
\end{equation}
where $n$ represents the number of $\bm{P}_\textit{i}$ in the set $\bm{F}$.

\subsubsection{Semantic region aware frontier point evaluation function} 
After obtaining $\bm{F}$, conventional methods evaluate each frontier point to determine the next best viewpoint goal $\bm{P}_{\textit{nbv}}$, by considering the information gain $\bm{G}(\bm{P}_\textit{i})$ and path cost $\bm{C}(\bm{P}_\textit{i})$. Differently, we propose the semantic region aware frontier evaluation function that also takes the semantic region information into consideration.

$\bm{G}(\bm{P}_\textit{i})$  evaluates the areas of unknown and obstacle regions in a square around the frontier point $\bm{P}_\textit{i}$, defined as follows.
\begin{equation}
\begin{aligned}
& \bm{G}(\bm{P}_\textit{i}) = f_s(g_{\textit{unkn}}) - f_s(g_{\textit{obs}}),\\
& g_{\textit{unkn}} \in \bm{S}_{\textit{unkn}}~,~g_{\textit{obs}} \in \bm{S}_{\textit{obs}}
\end{aligned}
\end{equation}
where $g_{\textit{unkn}}$ denotes the unknown region in the square and $g_{\textit{obs}}$ denotes the area of obstacle region in the square, and $f_s()$ is the function to compute the areas of $g_{\textit{unkn}}$ and $g_{\textit{obs}}$.

$\bm{C}(\bm{P}_{\textit{i}})$ evaluates the distance between $\bm{L}$ and $\bm{P}_\textit{i}$:
\begin{equation}
\begin{aligned}
\bm{C(\bm{P}_\textit{i}}) = \lVert \bm{P_\textit{i}} - \bm{L}\rVert
\end{aligned}
\end{equation}
 
In our semantic region aware frontier evaluation function, the semantic region information is also considered.
If $\bm{P}_\textit{i}$ and the robot are located in the same region (\textit{flag}=1), a positive reward $\bm{A}(\bm{P}_\textit{i})$ is added to the evaluation function. Otherwise, a negative reward $\bm{A}(\bm{P}_\textit{i})$ is added to the evaluation function. This mechanism encourages the robot to fully explore the current region before moving to the other region, which is formulated as follow. 
\begin{equation}
\mathbf{\mathit{S}}(\bm{P}_\textit{i})=
\left\{
\begin{aligned}
    &  \bm{\omega}_\textit{g} \cdot \bm{G}(\bm{P}_\textit{i}) - \bm{\omega}_\textit{c} \cdot \bm{C}(\bm{P}_\textit{i}) + \bm{A}(\bm{P}_\textit{i}) \: ,\:\:\textit{flag} = 1 \\
    &  \bm{\omega}_\textit{g} \cdot \bm{G}(\bm{P}_\textit{i}) - \bm{\omega}_\textit{c} \cdot \bm{C}(\bm{P}_\textit{i}) - \bm{A}(\bm{P}_\textit{i}) \: ,\:\:\textit{flag} = 0
\end{aligned}
\right.
\end{equation}
where $\bm{\omega}_\textit{g}$, $\bm{\omega}_\textit{c}$ are the weights of $\bm{G}(\bm{P}_\textit{i})$ and $\bm{C}(\bm{P}_\textit{i})$, respectively. $\bm{A}(\bm{P}_\textit{i})$ is set as an experimental value. $\mathbf{\mathit{S}}(\bm{P}_\textit{i})$ denotes the score of $\bm{P}_\textit{i}$. 

The frontier point with the highest score is selected as the next best viewpoint goal $\bm{P}_{\textit{nbv}}$ to perform autonomous navigation. 


\subsection{Multi-Type Map Construction}

When a mobile robot navigates autonomously, multi-type maps are constructed at the same time. The main challenges are aligning the coordinates of multi-type maps and harmonizing the computation threads of multi-type map construction.

2D occupancy grid map $\bm{M}_{\textit{occ}}$ is constructed by Cartographer SLAM \cite{r1} using laser scan and IMU data.
To guarantee the coordinate consistency of maps, we construct other types of maps using the reference coordinate of $\bm{M}_{\textit{occ}}$. 

Topological map and semantic map generation are based on the image map, which is converted from $\bm{M}_{\textit{occ}}$ using \textbf{Algorithm 1}. The first step is to create a matrix $\bm{M}_{\textit{img}}$. Next, the grids in $\bm{M}_{\textit{occ}}$ are traversed to judge whether they belong to $\bm{S}_{\textit{unkn}}$, $\bm{S}_{\textit{free}}$ or $\bm{S}_{\textit{obs}}$ ($\bm{S}_{\textit{know}}$ is classified as obstacle space $\bm{S}_{\textit{obs}}$ where the robot can not move due to the existing obstacles and unoccupied free space $\bm{S}_{\textit{free}}$). The corresponding pixels in $\bm{M}_{\textit{img}}$ are then respectively set to grey, white, and black. 

 \vspace{-10pt}
\begin{algorithm}[ht]
    \SetAlgoLined 
	\caption{Occupancy grid to image map}
	\KwIn{$\bm{M}_{\textit{occ}}$}
	\KwOut{$\bm{M}_{\textit{img}}$}
        Create a matrix $\bm{M}_{\textit{img}}$ $\gets$ height$(\bm{M}_{\textit{occ}})$, width$(\bm{M}_{\textit{occ}})$\\
	\For{i=1 to height$(\bm{M}_{\textit{occ}})$}{
           \For{j=1 to width$(\bm{M}_{\textit{occ}})$}{
               $g(i,j)$ $\gets$ \: $\bm{M}_{\textit{occ}}(i,j)$ \\
               \If{$g(i,j)$ $\in$ $\bm{S}_{\textit{unkn}}$ }{
			    $\bm{M}_{\textit{img}}(i,j)$ $\gets$ grey;
           }
               \If{$g(i,j)$ $\in$ $\bm{S}_{\textit{free}}$ }{
			    $\bm{M}_{\textit{img}}(i,j)$ $\gets$ white;
           }
               \If{$g(i,j)$ $\in$ $\bm{S}_{\textit{obs}}$ }{
			    $\bm{M}_{\textit{img}}(i,j)$ $\gets$ black;
           }
	}
}
\end{algorithm}
  \vspace{-12pt}
  
In topological map construction, to reduce measurement noise, we firstly apply binarization and morphological opening to filter out noise points. Then, the skeleton of topological map is extracted by thinning $\bm{M}_{\textit{img}}$ \cite{r39} in a traversing manner with 3$\times$3 matrix. The extracted skeleton elements are set to 1, and other elements are set to 0.

\begin{figure}[b]
\vspace{-15pt}
  \centering
  \includegraphics[width=0.42\textwidth]{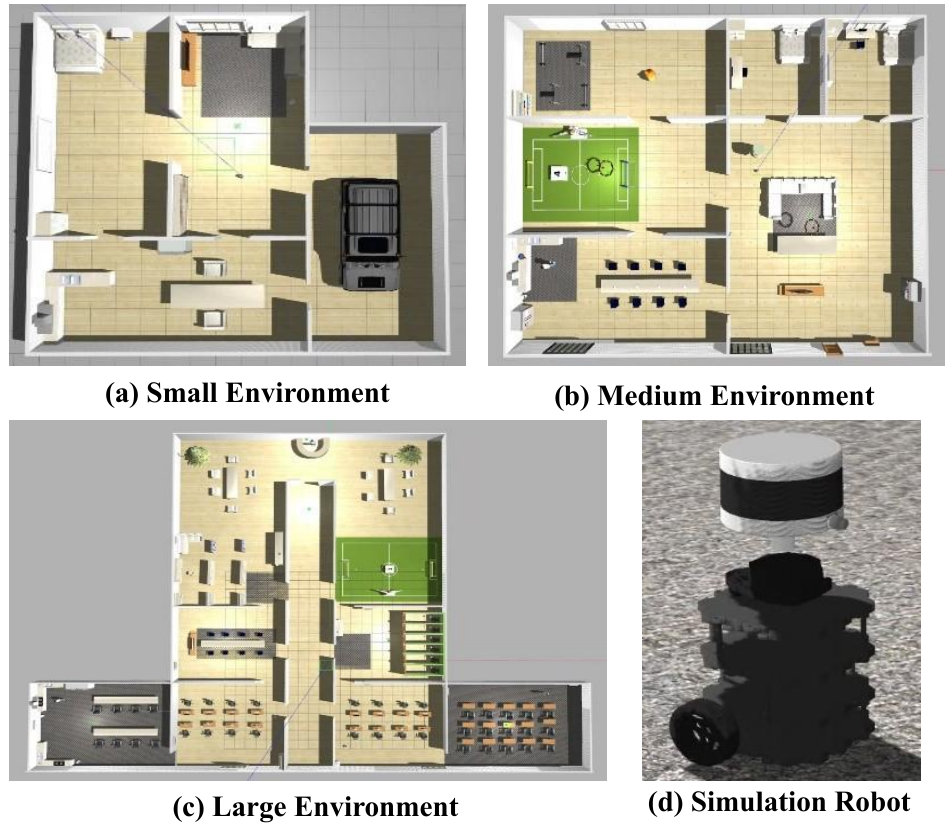}
  \caption{Simulation environment and simulation robot.} 
  \label{fig:simulation experiment environment}
    \vspace{-10pt}
\end{figure} 

In semantic map construction, referring to the processing of ROSE$^2$ \cite{r23}, the procedure involves: 1) extracting structural features and wall lines of the environment from $\bm{M}_{\textit{img}}$, 2) reconstructing geometric shapes of regions based on these features, and 3) assigning semantic ID and color to each region. Since the four types of maps share a common coordinate system, it is sufficient to obtain the robot's location in the semantic map to identify the semantic ID of the current region.
We then query and provide the frontiers and $\bm{P}_\textit{i}$ of that region to the semantic region aware autonomous navigation module to generate $\bm{P}_\textit{sem}$ and evaluate $\bm{P}_\textit{i}$.

\begin{figure*}[ht]
  \centering
  \includegraphics[width=1\textwidth]{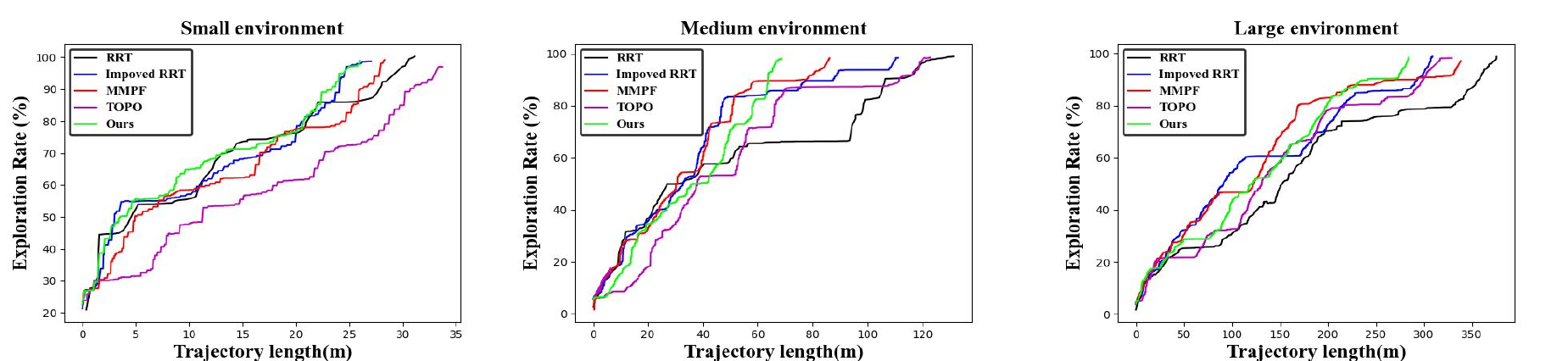}
    \vspace{-10pt} 
  \caption{Growing trend of exploration rate corresponding to
the increasing exploration trajectory length.} 
  \label{fig:Plot of exploration rate versus} 
    \vspace{-5pt} 
\end{figure*}
\begin{figure*}[h]
  \centering
  \includegraphics[width=1\textwidth]{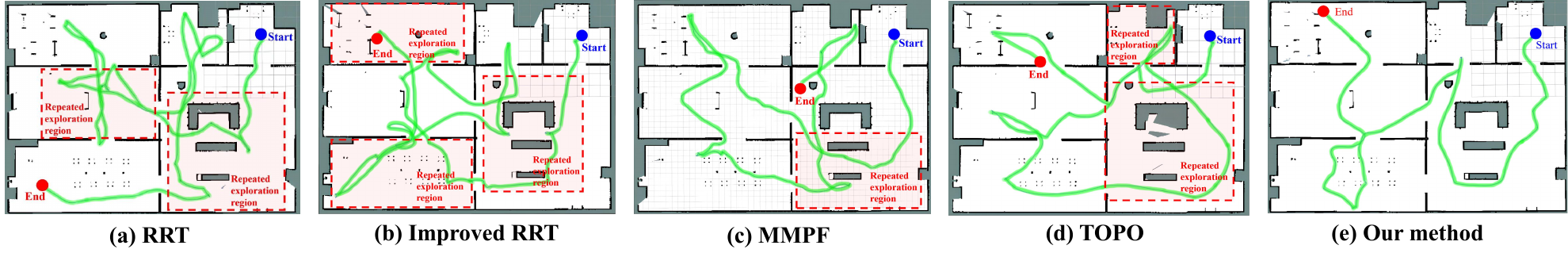}
    \vspace{-15pt}
  \caption{Comparison of exploration trajectories. Red boxes indicate repeated exploration region.} 
  \label{fig:Comparison of the exploration}
  \vspace{-20pt}
\end{figure*}

3D point cloud map is constructed using LIO-SAM \cite{r25} based on IMU data and point cloud data .

\section{ Experiment }
\vspace{-5pt}
\subsection{Settings}
\vspace{-2pt}

\subsubsection{Simulation environments and robot}
We set up three simulation environments using Gazebo \cite{r36}, including a small house ($187m^2$), a medium house ($450m^2$), and a large office ($1160m^2$), as shown in Fig.~\ref{fig:simulation experiment environment}. The Turtlebot3 Burger robot is used as the simulation robot, which is equipped with a $360^o$ laser scanner and a velodyne VLP-16 Lidar.

\subsubsection{Metrics}
The performance of autonomous exploration is evaluated by three metrics, including exploration time (i.e., the time consumption for exploring the whole environment), exploration trajectory length (i.e., the length of exploration trajectory for exploring the whole environment), and exploration rate (i.e., the ratio of the explored region to the whole environment).

\subsubsection{Baselines}

Four baseline methods, includes original RRT \cite{r6}, TOPO \cite{r28}, improved RRT and MMPF ( proposed in \cite{r22} ), are used in the experiments. 
RRT based methods are classical and commonly-used in autonomous exploration. MMPF \cite{r22}, and TOPO \cite{r28} are recently-proposed methods with publicly-available codes.

\begin{table}[ht] 
 \vspace{-5pt}
  \centering
  \renewcommand{\arraystretch}{1.2} 
  \caption{Exploration time and trajectory length comparison. T: Exploration time (s), L: Trajectory length (m). The best result is in bold.}
  \begin{tabular*}{\columnwidth}{@{\extracolsep{\fill}}lccccccc}
    \midrule
    \hspace{0.5cm} \multirow{2}* {Methods} & \multicolumn{2}{c}{Small} & \multicolumn{2}{c}{Medium} & \multicolumn{2}{c}{Large}\\
    \cmidrule(lr){2-7}
    & T & L & T & L  & T & L \\
    \midrule
    \hspace{0.1cm} RRT \cite{r6} & 171& 31& 593& 131& 1838& 375\\
    \hspace{0.1cm} Improved RRT \cite{r22} & 162& 27& 436& 112& 1282& 306\\
    \hspace{0.1cm} MMPF \cite{r22} & 125& 28& 350& 87& 1267& 338\\
    \hspace{0.1cm} TOPO \cite{r28} & \textbf{118} &33 &414 & 119 &1054 & 329\\
    \hline
    \hspace{0.1cm} Ours& 126 & \textbf{25} & \textbf{292} & \textbf{68} & \textbf{1018} & \textbf{283}\\
    \bottomrule
  \end{tabular*}
  \label{tab:time and length}
    \vspace{-15pt}
\end{table}


\subsection{Autonomous Exploration Comparison and Analysis}

\subsubsection{Exploration rate and trajectory length}
Fig.~\ref{fig:Plot of exploration rate versus} shows the growing trend of exploration rate corresponding to the increasing exploration trajectory length in small, medium, and large environments. We define that if exploration rate reaches 98\(\%\), an environment is supposed to be fully explored. 
We can observe from Fig.~\ref{fig:Plot of exploration rate versus} that all methods can fully explore the environment if exploration trajectory length is not limited. However, other methods ask for 
longer exploration trajectory to achieve full environment exploration, especially in medium and large environments that are bigger and more complex. In addition, in medium and large environment, there exist cases for other methods that exploration rate is not changing even though the length of the trajectory is increasing, implying that the robot is repeatedly moving in the previously-explored region. Instead, these cases are not frequently happened for our method. 

\subsubsection{Exploration time and trajectory length} 

We conducted the experiments to compare the exploration time and trajectory length of our method and baselines, and results in small, medium, and large environments are reported in Tab.~\ref{tab:time and length}. 
Our method asks for the least exploration time and the shortest exploration trajectory length in both medium and large environments.
For example, compared to classical RRT, our method achieves \textbf{50.7\%} exploration time reduction and \textbf{48.1\%} exploration trajectory length reduction in the medium environment. Based on the basic logic of our method, exploration time and trajectory will be further reduced with the increasing of complexity and scale of environments.   
Due to the simplicity of the small environment, the advantage of our method is not fully exhibited in the small environment. 

\begin{figure*}[ht]
  \vspace{3pt}
  \centering	
  \includegraphics[width=1\textwidth]{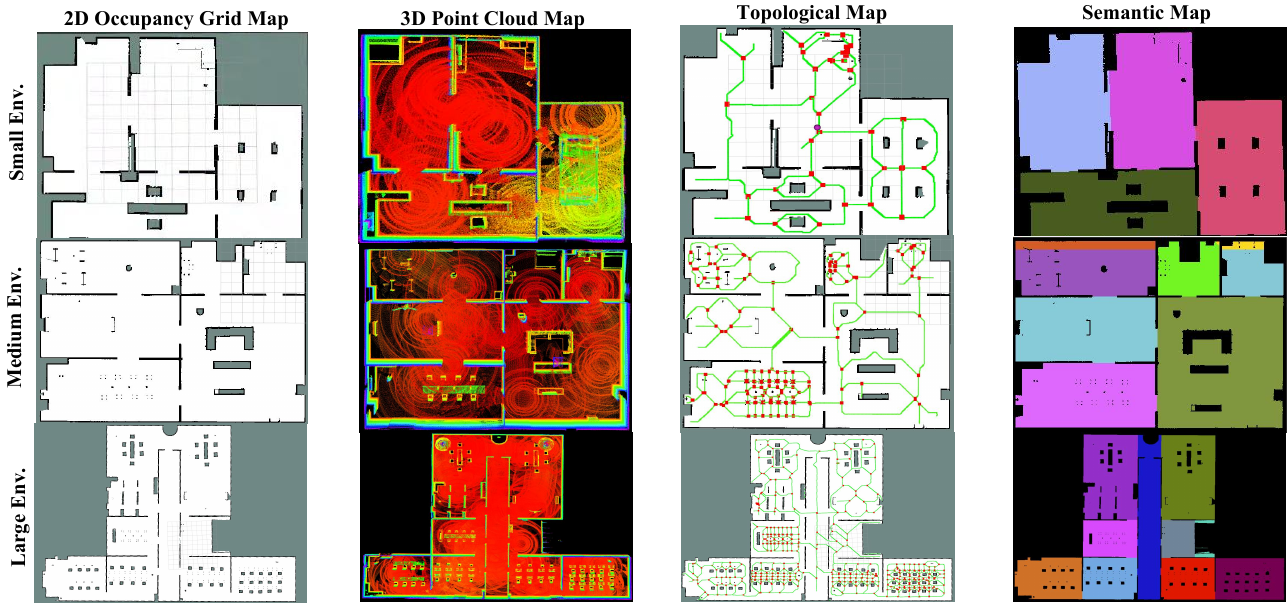}
  \caption{Four types of maps constructed by our semantic region aware autonomous exploration method. 
} 
  \label{fig:out put maps_all}
  \vspace{-18pt}
\end{figure*}

For further analysis, as shown in Fig.~\ref{fig:Comparison of the exploration}, we illustrated the exploration trajectories of different methods in the medium environment. The repeatly-explored regions are denoted by the red boxes. We can observe that other method make the robot enter and exit the same region more than one time to achieve the full exploration, while our method only asks for the robot to explore a region once, which significantly reduces the exploration time and trajectory length.

\subsection{Muti-Type Map Construction Comparison and Analysis}

Richer types of maps could support a wider range of downstream tasks and applications.
However, after reviewing the existing works in recent years, we find 1-3 types of maps are constructed, as summarized in Tab.~\ref{tab:table 4}. As far as we know, our method is the first to simultaneously construct four types of maps. Fig.~\ref{fig:out put maps_all} illustrates the muti-map construction results of our method in three environments. 
With the increasing of map types, many factors (e.g., computation thread conflict, map coordinate alignment, map storage, and map updating frequency) need to be taken into consideration. Computation thread conflict and map coordinate alignment have been well handled in our method. In the following, we analyze the update time and storage size of maps.

\begin{table}[ht]
  \centering
  \renewcommand{\arraystretch}{1.3} 
  \vspace{-10pt}
  \caption{The types of maps constructed by previous methods and our method. M1-M4 represents different types of maps. M1: Occupancy grid map, M2: Point cloud map, M3: Topological map, M4: Semantic map.}
  \begin{tabular*}{\columnwidth}{@{\extracolsep{\fill}}lcccc}
    \hline
    \hspace{1.5cm}Methods &  M1  &  M2  & M3  & M4 \\
    \hline
     \hspace{0.3cm}Cao et al. \cite{r48}\fontsize{6.0pt}{\baselineskip}\selectfont ICRA'2021 & & \checkmark   \\
     \hspace{0.3cm}Lehner et al. \cite{r49}\fontsize{6.0pt}{\baselineskip}\selectfont ICRA'2021 &\checkmark &   \\ 
      \hspace{0.3cm}Tian et al. \cite{r40}\fontsize{6.0pt}{\baselineskip}\selectfont ICRAI'2023  & \checkmark & &  \\
     \hspace{0.3cm}Gomez et al. \cite{r14}\fontsize{6.0pt}{\baselineskip}\selectfont ICRA'2020 & & \checkmark & \checkmark  \\
    \hspace{0.3cm}Zhang et al. \cite{r28}\fontsize{6.0pt}{\baselineskip}\selectfont RAL'2022  & \checkmark & & \checkmark \\
    \hspace{0.3cm}Ishikawa et al. \cite{r29}\fontsize{6.0pt}{\baselineskip}\selectfont SMC'2023 & \checkmark & & & \checkmark \\
    \hspace{0.3cm}Liu et al. \cite{r27}\fontsize{6.0pt}{\baselineskip}\selectfont IROS'2022 & & \checkmark & \checkmark & \checkmark \\
    \hline
    \hspace{0.3cm}Ours &\checkmark &\checkmark &\checkmark &\checkmark\\
    
    \hline
  \end{tabular*}
  \label{tab:table 4}
    \vspace{-10pt}
\end{table} 


Tab.~\ref{tab:table 3} reports the detailed update time and storage size of each kind of map in three environments.
Average update time for 2D occupancy grid map and topology map stays around 1$s$ even in different simulation environments. The 3D point cloud map has the shortest update time, which fluctuates around 0.2$s$ to quickly match the point cloud in consecutive frames. The semantic map requires longer update time (2.2$s$ to 3.1$s$) in bigger environment. 
In practice, the exploration procedure is real-time under these update time conditions, the robot did not stop to wait the update of the certain map, and the constructed maps are not deformed.
For the map storage, we use different file formats to save the different types of maps: 2D occupancy grid map (PGM), 3D point cloud map (PCD), topological map (JPG), and semantic map (PNG). After the full exploration, the storage space of all maps is approximately 3.1MB for the small environment, 7.5MB for the medium environment, and 17.4MB for the large environment, respectively. Standard industrial computers can fulfill these storage needs.
\begin{table}[ht]
\vspace{-5pt}
  \centering
  \caption{Storage size and update time of multi-type maps. S : storage size (KB), U : update time (s).}
  \setlength{\tabcolsep}{4pt} 
  \renewcommand{\arraystretch}{1.2} 
  \begin{tabular}{p{2.8cm} *{6}{c}}
    \toprule
    \multirow{2}{2.8cm}{\centering Map types} & \multicolumn{2}{c}{Small } & \multicolumn{2}{c}{Medium } & \multicolumn{2}{c}{Large }\\
    \cmidrule(lr){2-7}
    & S & U & S & U & S & U \\
    \midrule
    \hspace{0.3cm}\raggedright 2D Occupancy grid & 509.7 & 0.9& 707.0 & 0.9& 798.7 & 1.0\\
    \hspace{0.3cm}\raggedright 3D Point cloud & 2604.3 & 0.2& 6867.3 & 0.2& 16862.5 & 0.2\\
    \hspace{0.3cm}\raggedright Topological & 23.5& 1.1& 52.1& 1.0& 136.4 & 1.0\\
    \hspace{0.3cm}\raggedright Semantic & 6.1 & 2.2& 8.6 & 2.6& 18.3 & 3.1\\
    \bottomrule
  \end{tabular}
  \label{tab:table 3}
  \vspace{-10pt}
\end{table}
\section{ Conclusion }
In this paper, we propose a semantic region aware autonomous exploration method, which is able to avoid excessively-repeated explorations for the same region. The multi-type map construction method allows to simultaneously construct four types of maps in unknown indoor environments. Experimental results demonstrate that our method not only improves exploration efficiency but also provide multi-type map construction. In the future, we plan to extend our method to outdoor scenarios.

\bibliographystyle{IEEEtran}
\bibliography{references}

\end{document}